\documentclass[conference]{IEEEtran}
\IEEEoverridecommandlockouts
\usepackage{cite}
\usepackage{amsmath,amssymb,amsfonts}
\usepackage{algorithmic}
\usepackage{graphicx}
\usepackage{textcomp}
\usepackage{xcolor}
\input{./Definitions.tex}
\usepackage{multirow}
\usepackage{subfigure}
\usepackage{array}
\usepackage{booktabs}
\def\BibTeX{{\rm B\kern-.05em{\sc i\kern-.025em b}\kern-.08em
    T\kern-.1667em\lower.7ex\hbox{E}\kern-.125emX}}
\begin{document}

\title{Deep Minimax Probability Machine}

\author{\IEEEauthorblockN{Lirong He}
	\IEEEauthorblockA{\textit{SMILE Lab, School of Computer Science and Engineering} \\
		\textit{University of Electronic Science and Technology of China}\\
	Chengdu, China \\
		lirong\_he@std.uestc.edu.cn}
	\and
	\IEEEauthorblockN{Ziyi Guo}
	\IEEEauthorblockA{\textit{Cloud and Smart Industries Group} \\
		\textit{Tencent}\\
		Guangzhou, China \\
		ziyiguo94@gmail.com }
	\and
	\IEEEauthorblockN{Kaizhu Huang}
	\IEEEauthorblockA{\textit{Department of EEE} \\
		\textit{Xi’an Jiaotong-Liverpool University}\\
		Suzhou, China \\
		Kaizhu.Huang@xjtlu.edu.cn}
	\and
	\IEEEauthorblockN{Zenglin Xu\textsuperscript{1,2}}
	\IEEEauthorblockA{\textsuperscript{1}\textit{SMILE Lab, School of Computer Science and Engineering} \\
		\textit{University of Electronic Science and Technology of China}\\
		Chengdu, China}
		\IEEEauthorblockA{\textsuperscript{2}\textit{Center for Artificial Intelligence} \\
		\textit{Peng Cheng Laboratory}\\
		Shenzhen, China \\
		zenglin@gmail.com}	
	\and
   }

\maketitle

\begin{abstract}
Deep neural networks enjoy a powerful representation and have proven effective in a number of applications. However, recent advances show that deep neural networks are vulnerable to adversarial attacks incurred by the so-called adversarial examples. 
Although the adversarial example is only slightly different from the input sample, the neural network classifies it as the wrong class. In order to alleviate this problem, we propose the Deep Minimax Probability Machine (DeepMPM), which applies MPM to deep neural networks in an end-to-end fashion. In a worst-case scenario, MPM tries to minimize an upper bound of misclassification probabilities, considering the global information (i.e., mean and covariance information of each class). DeepMPM can be more robust since it learns the worst-case bound on the probability of misclassification of future data. Experiments on two real-world datasets can achieve comparable classification performance with CNN, while can be more robust on adversarial attacks. 
\end{abstract}


\section{Introduction}
Deep neural networks (DNNs) are adept at learning effective representation and have demonstrated significant success in a wide variety of applications, such as image classification ~\cite{krizhevsky2012imagenet}, speech recognition ~\cite{hinton2012deep,saon2015ibm} and language translation ~\cite{kalchbrenner2013recurrent,sutskever2014sequence}. 

However, recent advances show that they are vulnerable to adversarial examples, which are augmented  samples  perturbed imperceptibly but able to mislead the predictions of the neural networks ~\cite{szegedy2013intriguing,goodfellow6572explaining,lyu2015unified}. The existence of this issue prevents us from applying them to security-related applications, for example, self-driving cars~\cite{bojarski2016end}. 

It has attracted a lot of research interests in improving the adversarial robustness of deep neural networks. ~\cite{maharaj2015improving} suggests reducing the dimensionality of input data. ~\cite{bradshaw2017adversarial} indicates that their model is robust to adversarial examples putting  Gaussian processes on top of deep neural networks. Furthermore, ~\cite{miyato2018virtual} developed the semi-supervised version of adversarial training called Virtual Adversarial Training (VAT) where the output distribution was smoothed with a regularization term.

\begin{figure}[t]
	\begin{centering}
		\includegraphics[width = 0.85 \linewidth]{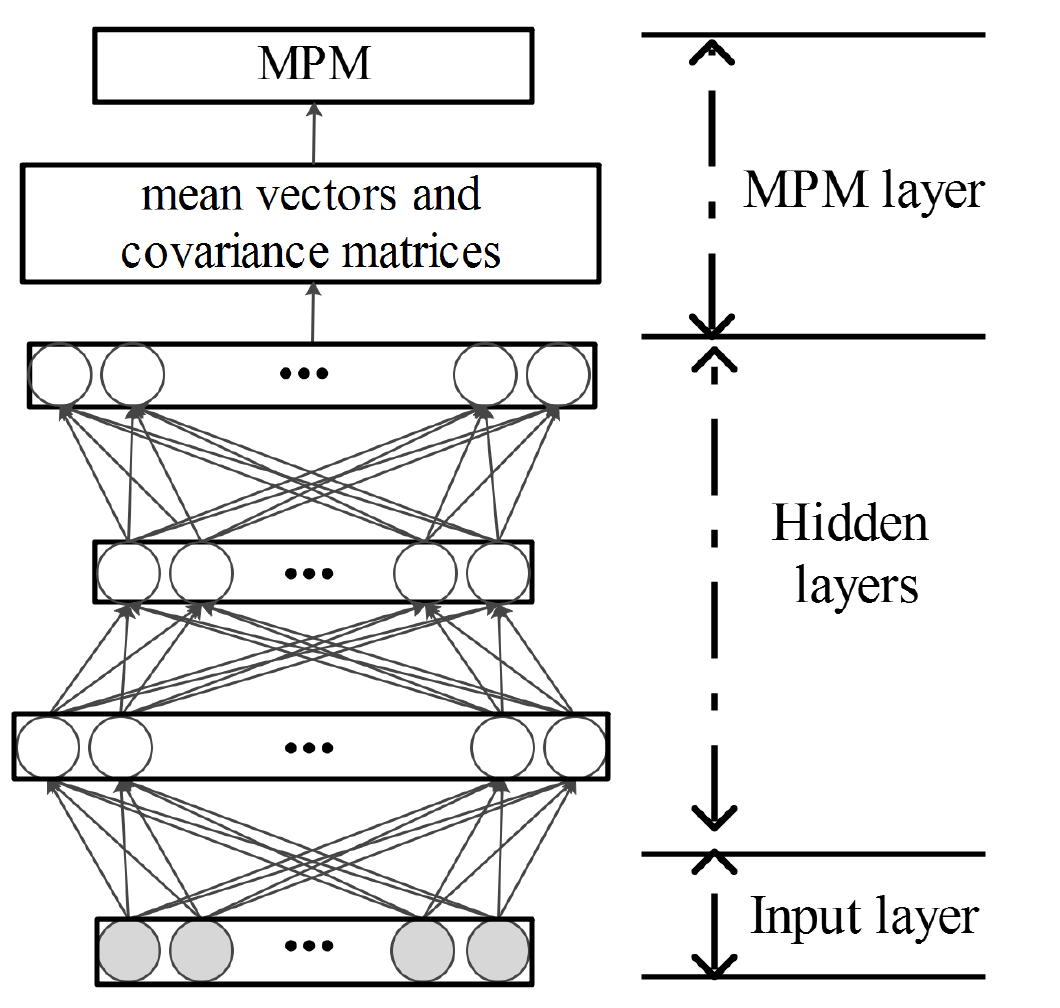}
		\caption{The graphical illustration of the DeepMPM model. The mainly difference of the CNN and the DeepMPM is the final layer: one uses the Softmax function, the other employs the MPM layer. This causes the two models to be different in the optimization section as well. And the DeepMPM obtains the means and covariance matrices of two classes through the powerful representation of the DNN.}
		\label{architecture}
	\end{centering}
\end{figure}

To alleviate such issues, we present an alternative model which equipts DNNs with the worst case misclassification bound provided by Minimax Probability Machine (MPM)~\cite{lanckriet2002minimax,huang2004minimum,huang2008maxi,huang2006imbalanced,xu2007feature}, leading Deep Minimax Probability Machine (DeepMPM). In essence, we put the MPM at the top of a deep neural network, as shown in Figure  ~\ref{architecture}.
Through exploring a powerful theorem\cite{Isii1962}, MPM can obtain the upper bound on the probability of misclassification for future data, i.e., the worst-case accuracy and hence leads to a robust classifier. In contrast, for classification, the traditional DNNs only consider the information of single example since they use softmax classifier. Combining MPM with DNNs could inherit the good advantages of both MPM and DNNs where robustness and accuracy would be integrated. In details, DeepMPM could take advantage of global information by introducing the global statistics of data, i.e., the mean and covariance of data, control misclassification probabilities robustly in the worst case for future data, and do well in learning effective hidden representation. In other words, applying MPM to DNNs can make up for the inadequacy of DNNs and are more robust to adversarial examples.  
Specifically, instead of maximizing the likelihood of labels for
data, we employ the objective function of MPM to promote our model to take into account global information. Importantly,  we engage the Lagrangian multiplier to optimize the DeepMPM which enables an end-to-end training fashion.

In order to evaluate the performance of the proposed model, we perform evaluations on two benchmark datasets, MINST and CIFAR-10, in two tasks including classification tasks and robustness to adversarial examples. Experimental results have demonstrated the encouraging performance. In particular, the significant improvement over the state-of-the-art methods has been observed in recognizing adversarial examples.

\section{Related Work}
\label{related}

It has been shown that deep neural networks are often vulnerable to adversarial examples. 
There is some literature on how to improve the adversarial robustness m from different perspectives. In the following, we present some literature on adversarial robustness.

The first aspect of work is based on adversarial training, specifically augmenting the training set with adversarial examples. It's said that adversarial training can improve the robustness of deep neural networks to adversarial examples and seems to hold the greatest promise.
\cite{goodfellow2014explaining,huang2015learning,ye2018learning,pan2019compressing,liu2018structured,tramer2017ensemble,madry2017towards}. For instance, in order to enhance the robustness of the neural networks, \cite{tramer2017ensemble} introduces adversarial training employing black-box attacks from similar networks. \cite{madry2017towards} demonstrates that if the perturbations calculated during training maximize the loss of the model,   adversarial training can be made robust to white-box attacks. The two methods described above are only for the specified attack method.

The second aspect of work is based on defensive distillation (i.e. transfer knowledge of complex networks to smaller networks) \cite{papernot2016distillation,papernot2017extending}. \cite{papernot2016distillation} employs knowledge of the network to train the networks to increase robustness. \cite{papernot2017extending} extends the defensive distillation measure by addressing the numerical instabilities encountered in \cite{papernot2016distillation}.

Meanwhile, there exists work to modify the input to improve the robustness \cite{maharaj2015improving,ross2018improving}.
 \cite{maharaj2015improving} points out that reducing the dimensionality of original data can improve the adversarial robustness of deep neural networks.
\cite{ross2018improving} shows that training neural networks with the input gradient regularization can enhance robustness to transferred adversarial examples created to fake out all of the other models.
In addition, the inherent characteristics of the model also affect robustness \cite{kurakin2016adversarial,guo2018sparse}. 
\cite{kurakin2016adversarial} finds that the robustness can be improved by increasing the capacity of the model. 
When this is used together with adversarial training, the effect is more pronounced.
\cite{guo2018sparse} demonstrates that sparse nonlinear deep neural networks are more robust than their corresponding dense networks. 

Deep neural networks can also combine the advantages of other methods to increase their own robustness. \cite{bradshaw2017adversarial} puts the Gaussian process at the top of a DNN, achieving good classification and robustness. In this paper, we combine the MPM with deep neural networks for the sake of their complementary strengths in robustness learning.

\section{Deep Minimax Probability Machine}
\label{sec:model}
In this section, we first provide an introduction to MPM. Further, we propose a novel model DeepMPM, which attempts to optimize an MPM optimization targeting in an end-to-end DNN fashion.

\subsection{Minimax Probability Machine}
Minimax Probability Machine (MPM) is proposed in~\cite{lanckriet2002minimax,huang2004minimum,huang2008maxi,huang2006imbalanced,xu2007feature}, which tries to minimize the upper bound of the probability of misclassification of future data in a worst-case setting. No assumptions with respect to the data distribution are required in MPM, while those assumptions lack generality and are often invalid.

Let $\x$ and $\y$ denote random vectors with mean vectors and covariance matrices given by $\x \sim (\bar{\x},\Sigma_{\x})$ and $\y \sim (\bar{\y},\Sigma_{\y})$ respectively in a binary classification problem, where $\x, \bar{\x},\y, \bar{\y} \in \mathbb{R}^{n}$ and $\Sigma_{\x},\Sigma_{\y} \in \mathbb{R}^{n \times n}$.

MPM seeks to determine the hyperplane ${\a}^{\top}\z=b$ ($\a,\z \in \mathbb{R}^{n}$ and $b \in \mathbb{R}$) which separates the two classes of data with maximal probability. The form of the MPM model is as follows:
\begin{align}
\max \limits_{\alpha,\a,b} \alpha \hspace{2ex} \text{s.t.}\hspace{1ex} &\inf \text{Pr}\{{\a}^{\top}\x \geq b\} \geq \alpha,\\
&\inf \text{Pr}\{{\a}^{\top}\y \leq b\} \geq \alpha. \nonumber
\end{align}
where $\alpha$ denotes the worst-case accuracy of the future data. Through a powerful theorem due to Isii\cite{Isii1962}, as extended by \cite{bertsimas2000moment}, this finally can be transformed into a convex optimization problem, as follows,
\begin{align}
\min \limits_{\a} \sqrt{\a^{\top}\Sigma_{\x}\a}+\sqrt{\a^{\top}\Sigma_{\y}\a} \hspace{2ex} \text{s.t.}\hspace{1ex} \a^{\top}(\bar{\x}-\bar{\y})=1,
\end{align}
or, equivalently 
\begin{align}
\min \limits_{\a} \lVert \Sigma_{\x}^{1/2}\a \rVert+ \lVert \Sigma_{\y}^{1/2}\a \rVert
\hspace{2ex} \text{s.t.}\hspace{1ex} \a^{\top}(\bar{\x}-\bar{\y})=1.
\end{align}
More specifically, this optimization problem is a second order cone program problem \cite{nesterov1994interior}. 
After obtaining the optimal solution $a_{\ast},b_{\ast}$, for a new data point $\z$, if $a_{\ast}\z \geq b_{\ast}$, $\z$ is classified as the class $\x$, otherwise $\z$ belongs to the class $\y$. In the following sections we will introduce DeepMPM.

\subsection{DeepMPM}
It is known that deep neural networks offer a powerful representation mechanism, which could learn adaptive basis functions focusing on local useful information. At the same time, MPM can directly minimize the maximum probability of misclassification with mean vectors and covariance matrices of the data considering the global structural information.

Therefore, we combine the MPM with deep neural networks for the sake of their complementary strengths in the classification task and robustness learning, namely Deep Minimax Probability Machine (DeepMPM). Specifically, we can interpret our model as applying the MPM to the final hidden layer of a deep neural network, instead of using softmax, as shown in Figure \ref{architecture}.

Let $g(\x,\w)$ denotes a nonlinear mapping given by a deep neural network, parametrized by weight $\w$. It can be said that through a neural network, we obtain effective representation for two classes of data, $g(\x,\w)$ and $g(\y,\w)$ respectively, making mean vectors and covariance matrices reliable. In details, 
\begin{align}
\x \sim (\bar{\x},\Sigma_{\x}) \rightarrow  g(\x,\w) \sim (\overline{g(\x,\w)},\Sigma_{g(\x,\w)}), \\
\y \sim (\bar{\y},\Sigma_{\y}) \rightarrow  g(\y,\w) \sim (\overline{g(\y,\w)},\Sigma_{g(\y,\w)}),
\end{align}
where $\overline{g(\x,\w)},\overline{g(\y,\w)}$ denote mean vectors of two classes of data respectively, and  $\Sigma_{g(\x,\w)}, \Sigma_{g(\y,\w)}$ denote covariance matrices of two classes of data respectively. For simplicity, we omit the parameter $\w$ of $g(\cdot)$ in the following sections. 

We desire a hyperplane $\a^{\top} g(\z)=b$ that separates the two classes of data points with maximal probability given the means and covariance matrices obtaining by a deep neural network. The formulation of our model is written as follows:
\begin{align}
\max \limits_{\alpha,\a,b} \alpha \hspace{2ex} \text{s.t.}\hspace{1ex} &\inf \text{Pr}\{{\a}^{\top}g(\x) \geq b\} \geq \alpha,\\
&\inf \text{Pr}\{{\a}^{\top}g(\y) \leq b\} \geq \alpha. \nonumber
\end{align}
With the powerful theorem used by the MPM ~\cite{Isii1962}, the optimized objective function of our model becomes,
\begin{align}
\min \limits_{\a} \hspace{1ex} &\sqrt{\a^{\top}\Sigma_{g(\x)}\a}+\sqrt{\a^{\top}\Sigma_{g(\y)}\a},  \\
&\text{s.t.}\hspace{1ex} \a^{\top}(\overline{g(\x)}-\overline{g(\y)})=1. \nonumber
\end{align}
In order to train our model in an end-to-end fashion, we employ the Lagrangian multiplier method to perform optimization. With the introduction of a Lagrange multiplier $\lambda$, we can minimize the objective function,
\begin{align}
\mathcal{L}=\sqrt{\a^{\top}\Sigma_{g(\x)}\a}+\sqrt{\a^{\top}\Sigma_{g(\y)}\a}+\lambda(\a^{\top}(\overline{g(\x)}-\overline{g(\y)})-1).
\label{loss}
\end{align}

We would like to underline that we jointly learn all our model parameters, $\phi =\{\w, \a \}$,  including $\w$ the weights of the neural network, and $\a$ the parameter of the hyperplane for the MPM. In addition, we train DeepMPM with back propagation in an end-to-end fashion, using the chain rule to calculate derivatives about all the parameters. Particularly, the derivative of weight $w$ is written as,
\begin{align}
\frac{\partial \mathcal{L}}{\partial \w}=\frac{\partial \mathcal{L}}{\partial \Sigma_{g(\x)}} \frac{\partial \Sigma_{g(\x)}}{\partial g(\x)} \frac{\partial g(\x)}{\partial \w}+ \frac{\partial \mathcal{L}}{\partial \Sigma_{g(\y)}} \frac{\partial \Sigma_{g(\y)}}{\partial g(\y)} \frac{\partial g(\y)}{\partial \w} \nonumber \\
+ \frac{\partial \mathcal{L}}{\partial \overline{g(\x)}} \frac{\partial \overline{g(\x)}}{\partial g(\x)} \frac{\partial g(\x)}{\partial \w} + \frac{\partial \mathcal{L}}{\partial \overline{g(\y)}} \frac{\partial \overline{g(\y)}}{\partial g(\y)} \frac{\partial g(\y)}{\partial \w}.
\end{align}

We apply the Nesterov momentum version of mini-batch SGD to optimize our model. Related methods have shown that mini-batch learning of distribution parameters (in detail, mean vectors and covariance matrices) is feasible if the batch size is adequately large~\cite{wang2015deep,wang2015unsupervised}.

With the optimized parameters $\phi_{\ast}=\{ \w_{\ast}, \a_{\ast} \}$, we can obtain $b_{\ast}$ and $\alpha^{\ast}$ respectively, as follows,
\begin{align}
b_{\ast}=\a_{\ast}^{\top}\overline{g(\x)}-\frac{\sqrt{\a_{\ast}^{\top}\Sigma_{g(\x)}\a_{\ast}}}{\sqrt{\a_{\ast}^{\top}\Sigma_{g(\x)}\a_{\ast}}+\sqrt{\a_{\ast}^{\top}\Sigma_{g(\w)}\a_{\ast}}}.
\end{align}
and 
\begin{align}
	\alpha^{\ast}=\frac{1}{(\sqrt{\a^{\ast^{\top}}\Sigma_{g(\x)}\a^{\ast}}+\sqrt{\a^{\ast^{\top}}\Sigma_{g(\y)}\a^{\ast}})^2+1}.
\end{align}

$\alpha^{\ast}$ represents the worst-case accuracy of the future data.
In general, traditional machine learning is fully data-driven, with the goal of maximizing the accuracy of the known data  in the average sense, while our model is to maximize the accuracy of the future data in the worst sense, which is more robust.

The label of a test instance can be given by $\mathrm{sign}(\a_{\ast}^{\top} g(\z,\w_{\ast})- b_{\ast})$: if the formulation is +1, $\z$ is classified as the class $\x$, otherwise $\z$ is classified as the class $\y$.

For multi-class classification, one can train a neural network with shared feature extraction and an integrated loss function via the one-vs-others or one-vs-one schemes, and then compare the $\alpha$ values for each class. Since the goal of this paper is to provide a seminal work of extending DNNs with MPM to evaluate its robustness, we leave this extension as the future work.

\section{Experiment}
\label{section:exp}
\begin{figure*}[htb!]
	\begin{centering}
		\includegraphics[width = 0.98 \linewidth]{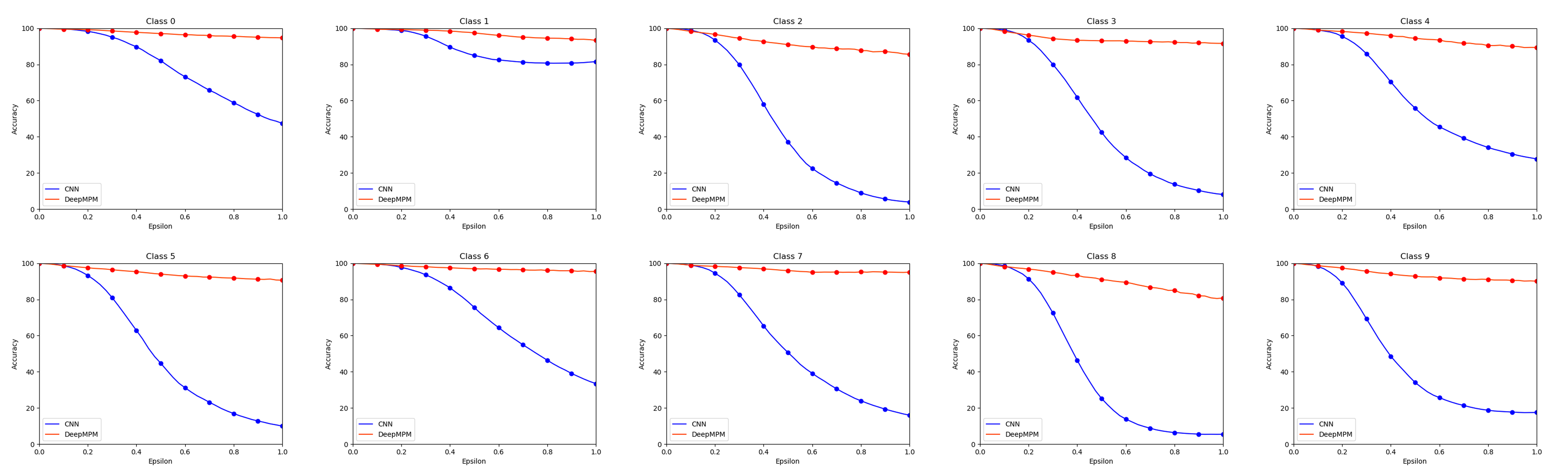}
		\caption{The accuracy of ten binary classifiers on MNIST for FGSM attacks. The horizontal axis represents the size of $\epsilon$. It's seen that the Accuracy of the DeepMPM decreases much more slowly with the size of adversarial perturbation for all ten classifiers. Thus, DeepMPM is more robust.}
		\label{acc_fgsm_minist}
	\end{centering}
\end{figure*}
\begin{figure*}[htb!]
	\begin{centering}
		\includegraphics[width = 1 \linewidth]{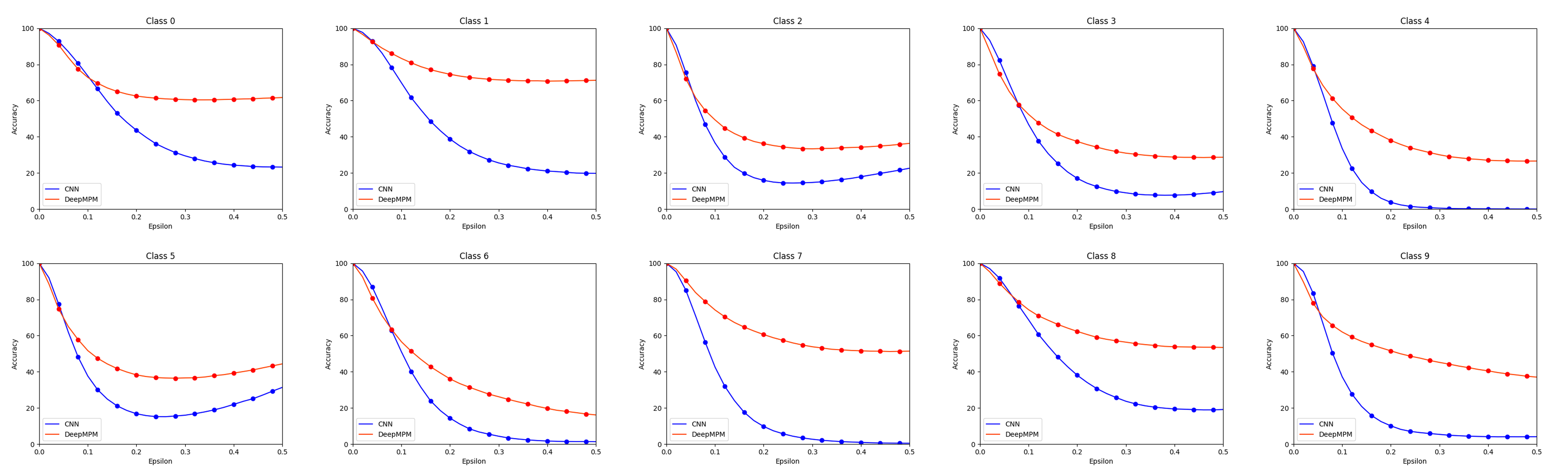}
		\caption{The accuracy of ten binary classifiers from the FGSM attacks on CIFAR-10. DeepMPM's
			accuracy decreases significantly slower with size of
			adversarial perturbation.}
		\label{acc_fgsm_cifar10}
	\end{centering}
\end{figure*}

We evaluate our model across multiple scenarios on two datasets  MNIST~\cite{lecun1998mnist} and CIFAR-10~\cite{krizhevsky2009learning} respectively. 
Specifically, we evaluate the performance of DeepMPM on classification tasks and compare the accuracy of two datasets in the first place. Then we examine the robustness of our model under the attacks of adversarial examples.

All experiments are performed on a Linux machine with eight 4.0GHz CPU cores and 32GB RAM, one GTX 1080 Ti GPU card with 12GB RAM. We implemented DNNs based on Pytorch, a general deep learning platform. 

\subsection{Classification}
In this subsection, we report the performance of our model on binary classification tasks. In particular, we test our model on MNIST and CIFAR-10 datasets. As MPM is designed to solve binary classification problems, we train ten binary classifiers with CNN and DeepMPM. For each binary classifier, we use the one-vs-others approach instead of the one-vs-one approach.

For the experiments on MNIST, we exploit the network architecture in the Pytorch basic MNIST example\footnote{available at https://github.com/pytorch/examples/tree/0.4/mnist}
as the base CNN,  which contains two convolutional layers and two fully connected layers, as shown in Table~\ref{archi_mnist}.

 \begin{table}[htbp]
 	\begin{center}
 		\begin{tabular}{cc}
 			\specialrule{0.05em}{1.5pt}{1.5pt}
 			CNN &  DeepMPM      \\ 
 			\specialrule{0.05em}{1pt}{1pt}
 			\multicolumn{2}{c}{Conv. (5 by 5, 10 channels)}\\
 			\multicolumn{2}{c}{Max pooling (2 by 2, padding is SAME)}  \\
 			\multicolumn{2}{c}{Conv. (5 by 5, 20 channels)}  \\
 			\multicolumn{2}{c}{Dropout}  \\
 			\multicolumn{2}{c}{Max pooling (2 by 2, padding is SAME)}   \\
 			\multicolumn{2}{c}{FC (to 50 units)}  \\
 			FC (to 2 units)  &  \multirow{2}{*}{MPM (to 1 unit)}  \\
 			Softmax   \\
 			\hline
 		\end{tabular}
 	\end{center}
 		\caption{
 		   The CNN and DeepMPM architectures we use on MNIST. The FC is the abbreviation of the fully connected layer.
 			}
 			\label{archi_mnist}
 \end{table}

While CIFAR-10 contains images of more complicated objects, we employ a deeper network. The base CNN architecture we use comes from~\cite{papernot2016distillation} containing four convolutional layers and two fully connected layers, as shown in Table~\ref{archi_cifar10}.

 \begin{table}[htbp]

 \begin{center}
 \begin{tabular}{cc}
\specialrule{0.05em}{1.5pt}{1.5pt}
 CNN &  DeepMPM      \\ 
\specialrule{0.05em}{1pt}{1pt}
      \multicolumn{2}{c}{Conv. (3 by 3, 64 channels)}\\
      \multicolumn{2}{c}{Conv. (3 by 3, 64 channels)}  \\
      \multicolumn{2}{c}{Max pooling (2 by 2, padding is SAME)}  \\
      \multicolumn{2}{c}{Conv. (3 by 3, 128 channels)}  \\
      \multicolumn{2}{c}{Conv. (3 by 3, 128 channels)}  \\
      \multicolumn{2}{c}{Max pooling (2 by 2, padding is SAME}   \\
      \multicolumn{2}{c}{FC (to 256 units)}  \\
      FC (to 2 units) & \multirow{2}{*}{MPM (to 1 unit)} \\ 
      Softmax \\ 
     \hline
     \end{tabular}
     \end{center}
     \begin{small}
     \caption{
     \small{The CNN and DeepMPM architectures we use on CIFAR-10.
 	}
 	\label{archi_cifar10}
 	}
 \end{small}
 \end{table}
The same training set is used for all the experiments. Particularly, each model is trained end-to-end for 100 epochs, and the learning rate decay would be performed on the 50th and 80th epoch, with the decay factor as $0.1$.
Hyper-parameters for CNN models and MPM models are set the same except for the learning rate and momentum. 
We choose the small learning rate and momentum empirically, since greater learning rates and larger momentums would produce Inf in gradient. Namely,  the learning rate and  momentum are set to $1e-2$ and $0.9$ respectively for CNN models, while they are set to $1e-3$  and $0.5$ for DeepMPM models. 

The accuracy of two comparison models on 10 tasks of MNIST and CIFAR-10 are reported in Table~\ref{binary_acc}. 
DeepMPM performs better than CNN on 4 out of 10 tasks for MNIST, and the largest gap between our model and CNN on other 6 tasks is $0.11\%$. As for tasks on CIFAR-10, the average gap between the accuracy of the two models is $0.31\%$.
From the results above, we could observe that our model could achieve comparable performance with CNN (the state-of-art method) on the ordinary classification tasks.

\begin{table}[htbp]
	\begin{center}
		\begin{tabular}{p{1cm}<{\centering}|cc|cc}
			\specialrule{0.05em}{2pt}{2pt}
			\multirow{2}{*}{}& \multicolumn{2}{c|}{MNIST} &  \multicolumn{2}{c}{CIFAR-10} \\
			& \ CNN & DeepMPM &  \ CNN & DeepMPM \\
			\specialrule{0.05em}{1.5pt}{1.5pt}
			0 & \ 99.86 & 99.78 & 95.95 & 95.55 \\
			\specialrule{0.05em}{1.5pt}{1.5pt}
			1 & \ 99.83 & 99.86 & 97.71 & 97.13 \\
            \specialrule{0.05em}{1.5pt}{1.5pt}
			2 & \ 99.76 & 99.69 & 93.95 & 94.10 \\
			\specialrule{0.05em}{1.5pt}{1.5pt}
			3 & \ 99.80 & 99.83 & 92.42 & 92.08 \\
			\specialrule{0.05em}{1.5pt}{1.5pt}
			4 & \ 99.87 & 99.80 & 94.55 & 94.37 \\
			\specialrule{0.05em}{1.5pt}{1.5pt}
			5 & \ 99.82 & 99.83 & 94.84 & 94.59\\
			\specialrule{0.05em}{1.5pt}{1.5pt}
			6 & \ 99.77 & 99.66 & 96.31 & 95.81 \\
			\specialrule{0.05em}{1.5pt}{1.5pt}
			7 & \ 99.72 & 99.70 & 96.48 & 96.18 \\
			\specialrule{0.05em}{1.5pt}{1.5pt}
			8 &\ 99.84 & 99.68 & 97.35 & 97.15 \\
			\specialrule{0.05em}{1.5pt}{1.5pt}
			9 & \ 99.57 & 99.60 & 97.43 & 96.96 \\
			\specialrule{0.05em}{1.5pt}{1.5pt}
		\end{tabular}
	\end{center}
		\caption{
			The accuracy of the ten binary classification tasks of MNIST and CIFAR-10 (\%). It shows that the DeepMPM model can achieve comparable performance with the CNN model. 
			\label{binary_acc}
		}
\end{table}
\begin{figure*}[htb!]
	\begin{centering}
		\includegraphics[width = 0.98 \linewidth]{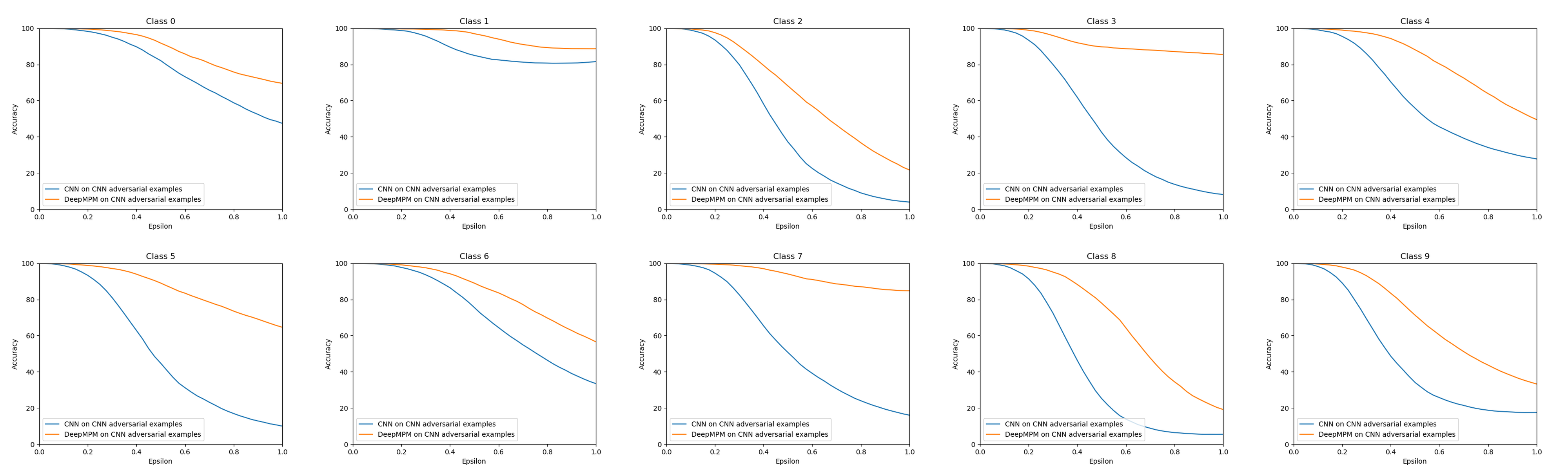}
		\caption{The accuracy of both models on MNIST when applying both models in the adversarial examples generated by CNN  for the FGSM attacks (\%). With the adversarial examples generated by the CNN, although the DeepMPM performs worse than the case when applied to its own adversarial examples, its accuracy is much higher than that of the CNN.}
		\label{acc_fgsm_minist_ON_CNN_Ad}
	\end{centering}
\end{figure*}

\begin{figure*}[hbt!]
	\begin{centering}
		\includegraphics[width = 0.98 \linewidth]{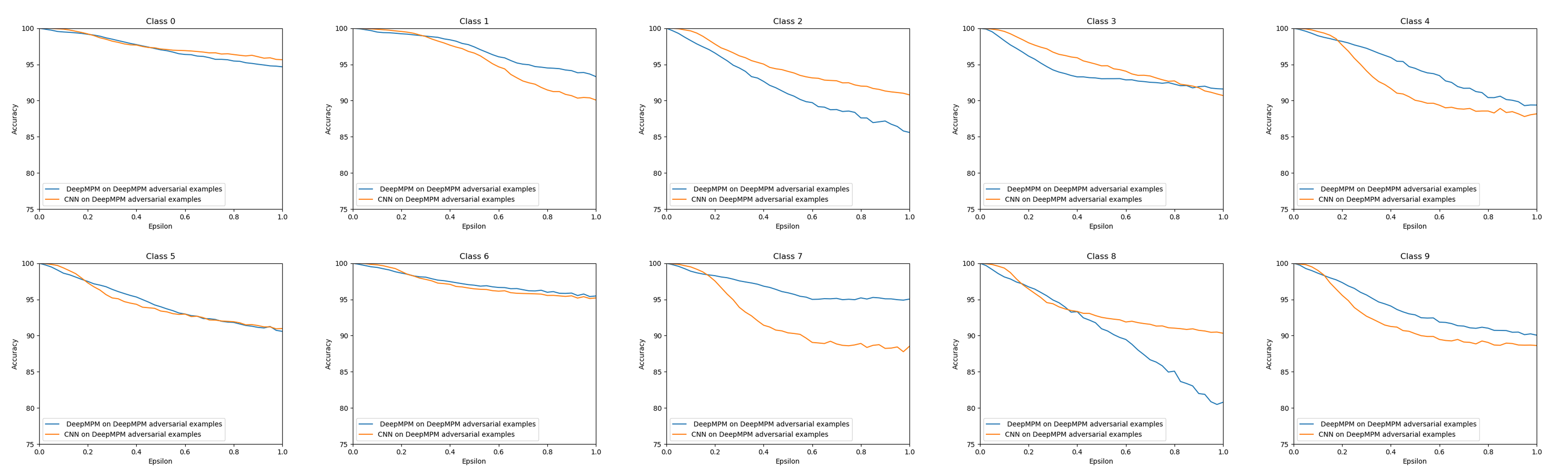}
		\caption{The accuracy of both models on MNIST  when applying both models in the adversarial examples generated by the DeepMPM for the FGSM attacks (\%). In order to see the change more clearly, the value of the vertical axis starts from 75\%. When attacking the CNN model on adversarial examples generated by the DeepMPM, the CNN has a great prediction on several binary classifications.}
		\label{acc_fgsm_minist_ON_MPM_Ad}
	\end{centering}
\end{figure*}
\subsection{Robustness to adversarial examples}
There is a large amount of literature on producing adversarial examples with different methods \cite{moosavi2017universal,goodfellow6572explaining,lyu2015unified,carlini2017towards} 
 
since \cite{szegedy2013intriguing} pointed out neural networks are always vulnerable to adversarial examples. Adversarial examples are generated data which have minor differences from original data yet can mislead the deep networks to give the incorrect prediction. 
Adversarial attacks can be divided into two categories, targeted attacks and non targeted attacks respectively. 
Targeted attacks are those trying to find a minimal permutation that leads the classifier to the desired class prediction (target), while non-targeted attacks only care about the case that the classifier gives a different answer. 
In this paper, we focus on non-targeted attacks.  It is noted that  non-targeted could actually be viewed as targeted if classification tasks are all binary.

To validate the robustness of our model, we test the base CNN and our model on two attacks: the Fast Gradient Sign Method (FGSM)~\cite{goodfellow6572explaining} and the L2 optimization attack of Carlini and Wagner~\cite{carlini2017towards}.

\subsubsection{The fast gradient sign method}
The fast gradient sign method (FGSM) ~\cite{goodfellow6572explaining} is a one shot method for generating adversarial examples and can be considered as one special case of generalized gradient regularized method later developed in~\cite{lyu2015unified}. It tries to find  the perturbation through the direction of the sign function of the gradient, which can be calculated by back-propagation. For a given image $\x$, the perturbation can be written as, 
\begin{align}
\bm{\eta}=\epsilon \hspace{0.5ex} \text{sign}(\nabla_{\x}J(\bm{\theta},\x)),
\end{align}
where $J(\bm{\theta},\x)$ represents the model loss function with model parameters, $\epsilon$ is a small scalar value that controls the magnitude of the perturbation and $\text{sign}(\cdot)$ denotes the sign function. Then the adversarial example $\x'$ is computed as: $\x'=\x+\bm{\eta}$.

For deep convolutional neural networks, the loss function is given as cross-entropy, namely the likelihood of the target class label. 
For our proposed DeepMPM model, the loss function is Equation~(\ref{loss}) from the original MPM model.
Attacks are performed on the models trained for classification tasks mentioned in the previous section.

To evaluate the robustness of our proposed model, we report the accuracy of classifiers on test sets of MNIST and CIFAR-10, with the magnitude factor $\epsilon$ ranges from small to large. 
For  the MNIST experiment, $\epsilon$ ranges from 0 to 1, with step size $0.025$. Meanwhile, in the CIFAR-10 experiment, we set the value of $\epsilon$'s from 0 to $0.5$, and gradually increase it by $0.02$. 

\begin{table}[htbp]
	\begin{center}
		\setlength{\tabcolsep}{1.4mm}{
			\begin{tabular}{p{0.4cm}<{\centering}l|cc|cc}
				\specialrule{0.05em}{2pt}{2pt}
				\multirow{2}{*}{}& \multirow{2}{*}{}& \multicolumn{2}{c|}{MNIST} &  \multicolumn{2}{c}{CIFAR-10} \\
				& & CNN & DeepMPM &   CNN & DeepMPM \\
				\specialrule{0.05em}{1.5pt}{1.5pt}
				\multirow{2}{*}{0} & CNN & \ 0.00 & 94.09 & 0.00 & 89.71 \\
				& DeepMPM & \ 99.99 & 89.50 & 98.16 & 0.00 \\
				\specialrule{0.05em}{1.5pt}{1.5pt}
				\multirow{2}{*}{1} & CNN & \ 0.00 & 99.66 & 0.00 & 95.35 \\
				& DeepMPM & \ 99.47 & 88.00 & 98.73 & 0.05 \\
				\specialrule{0.05em}{1.5pt}{1.5pt}
				\multirow{2}{*}{2} & CNN & \ 0.01 & 97.96 & 0.00 & 77.81 \\
				& DeepMPM & \ 99.98 &  78.95 & 96.15 &  0.01 \\
				\specialrule{0.05em}{1.5pt}{1.5pt}
				\multirow{2}{*}{3} & CNN & \ 0.00 & 98.74 & 0.00 & 77.81 \\
				& DeepMPM & \ 99.95 & 86.64 & 97.58 & 0.00 \\
			   \specialrule{0.05em}{1.5pt}{1.5pt}
				\multirow{2}{*}{4} & CNN & \ 0.13 & 98.50 & 0.00 & 85.65 \\
				& DeepMPM & \ 99.91 & 86.90 & 97.83 & 0.00 \\
				\specialrule{0.05em}{1.5pt}{1.5pt}
				\multirow{2}{*}{5} & CNN & \ 0.00 & 98.88 & 0.00 & 89.48 \\
				& DeepMPM & \ 99.93 & 89.00 & 97.28 & 0.00 \\
			    \specialrule{0.05em}{1.5pt}{1.5pt}
				\multirow{2}{*}{6} & CNN & \ 0.00 & 95.25 & 0.00 & 87.19 \\
				& DeepMPM & \ 99.97 & 89.93 & 97.96 &  0.01 \\
				\specialrule{0.05em}{1.5pt}{1.5pt}
				\multirow{2}{*}{7} & CNN & \ 0.00 & 99.65 & 0.00 & 95.40 \\
				& DeepMPM & \ 99.92 & 88.59 & 94.17 & 0.11 \\
				\specialrule{0.05em}{1.5pt}{1.5pt}
				\multirow{2}{*}{8} & CNN & \ 0.00 & 91.00 & 0.00 & 89.86 \\
				& DeepMPM & \ 99.89 & 88.14 & 98.40 & 0.03 \\
				\specialrule{0.05em}{1.5pt}{1.5pt}
				\multirow{2}{*}{9} & CNN & \ 0.00 & 97.85& 0.00 & 66.30 \\
				& DeepMPM & \ 99.89 & 87.93 & 97.77 & 0.22 \\
				\specialrule{0.05em}{1.5pt}{1.5pt}
		\end{tabular}}
	\end{center}
		\caption{
			The accuracy of the ten binary classification tasks of MNIST and CIFAR-10 over L2 C\&W attack examples(\%).
			\label{c&w_acc}
		}
\end{table}

We report the accuracy of the model under self-attack scenario in the first place, in Figure~\ref{acc_fgsm_minist} (MNIST), and Figure~\ref{acc_fgsm_cifar10} (CIFAR-10).
Note that attacks are performed among examples which are originally classified correctly. For the sake of fairness, only examples that are correctly classified by both models are tested. 
Therefore, at the starting point where $\epsilon$ is equal to 0, the accuracy of both models are 100$\%$.
For MNIST, in Figure~\ref{acc_fgsm_minist}, it can be observed that the accuracy of DeepMPM remains high while that of the CNN model reduces significantly as $\epsilon$ increases.  As for the harder CIFAR-10 tasks, in Figure~\ref{acc_fgsm_cifar10}, though the accuracy decreases are observed, DeepMPM still generates much better predictions than CNN.
The above results demonstrate the great robustness of DeepMPM under FGSM attacks in the self-attack scenarios.

To illustrate the performance of the FSGM attack examples transferred between different models, we plot the accuracy of classifiers when both models are applied to adversarial examples generated by the CNN model, as shown in Figure~\ref{acc_fgsm_minist_ON_CNN_Ad}. In this paper, We mainly present the experimental results on the MNIST dataset.
For the experiments on MNIST we find that in most classifiers, although the DeepMPM model performs worse than the case when applied to its own adversarial examples, its accuracy is observed much higher than the accuracy of CNN. 
This indicates that the performance of DeepMPM is better than that of CNN. 

We also show the accuracy of classifiers when two models are applied to adversarial examples generated by the DeepMPM model, shown in Figure~\ref{acc_fgsm_minist_ON_MPM_Ad}. Similarly, We mainly present the experimental results on the MNIST dataset.
The accuracy of CNN has increased a lot compared to applying to its own adversarial examples. And the accuracy of the two models is almost the same. 

From the Figure~\ref{acc_fgsm_minist_ON_CNN_Ad} and Figure~\ref{acc_fgsm_minist_ON_MPM_Ad}, it can be seen that the attack examples generated by the DeepMPM model are less aggressive,  while the attack examples generated by the CNN model are universal.

\subsubsection{L2 optimization attack of Carlini and Wagner}

To further validate the robustness of our model, we perform attacks with the L2 optimization attack of Carlini and Wagner \cite{carlini2017towards}. This method finds the perturbation by minimizing the following function,
\begin{align}
\Vert \bm{\eta}\Vert_{2}^{2}+c f(\x + \bm{\eta}).
\label{c&w_function}
\end{align}
where $c$ is a constant chosen by the binary search method and  $f(\cdot)$ we used is defined as below for non targeted attacks.
\begin{align}
f(\x')=Z(\x')_{l}-\max \{Z(\x')_{i}:i \neq l\}.
\label{Z_function}
\end{align}
$Z(\x')_{i}$ denotes the pre softmax predictions for the class $i$ on image $\x'$, and $l$ represents the correct class.

As no softmax function is carried out in our model, this attack might fail in generating adversarial examples directly from pre-softmax values like in CNN.
It is  known that we get the prediction by the value of $\a_{\ast}^{\top} g(\z,\w_{\ast})- b_{\ast}$: if the value is greater than or equal to 0, the new point $\z$ is classified as the class $\x$, otherwise $\z$ is classified as the class $\y$. 
We found that the value of the equation $S(\a_{\ast}^{\top} g(\z,\w_{\ast})- b_{\ast})$ could represent the probability that a sample belongs to class $\x$ to some extent, where $S(\cdot)$ denotes the sigmoid function, and the probability that a sample  belongs to class $\y$ can be obtained by the value of the equation $1-S(\a_{\ast}^{\top} g(\z,\w_{\ast})- b_{\ast})$. 
Therefore, we reversely calculate the pre-softmax values with the probabilities, which is the same as done in \cite{bradshaw2017adversarial}. 
Same as the implementation of \cite{carlini2017towards}, we actually optimize $\eta$ in a transformed space.

Different from the FGSM attack, L2 optimization adversarial attack is an iterative attack which optimizes over an example for multiple times.
Thus the optimized adversarial examples can easily reach rather a high attack success rate than FGSM generated examples. 

In Table ~\ref{c&w_acc}, we report the accuracy of models over the generated attack examples of L2 optimization of Carlini and Wagner.
The row represents the targeted model of attacks while the column represents the model be tested on.
As can be observed, DeepMPM can recognize attack examples with high confidence (87.36\% averagely) under self-attack scenario for MNIST while CNN got nearly 0\% accuracy. 
For Cifar-10, both models are perfectly attacked under self-attack examples, yet DeepMPM's accuracy suffers more on transfer attacks where examples of one model are tested on the other. (It's partly due to the universality of CNN targeted examples.)

\begin{figure}[htb!]
	\vspace{-1ex}
	\begin{centering}
		\includegraphics[width = 0.98 \linewidth]{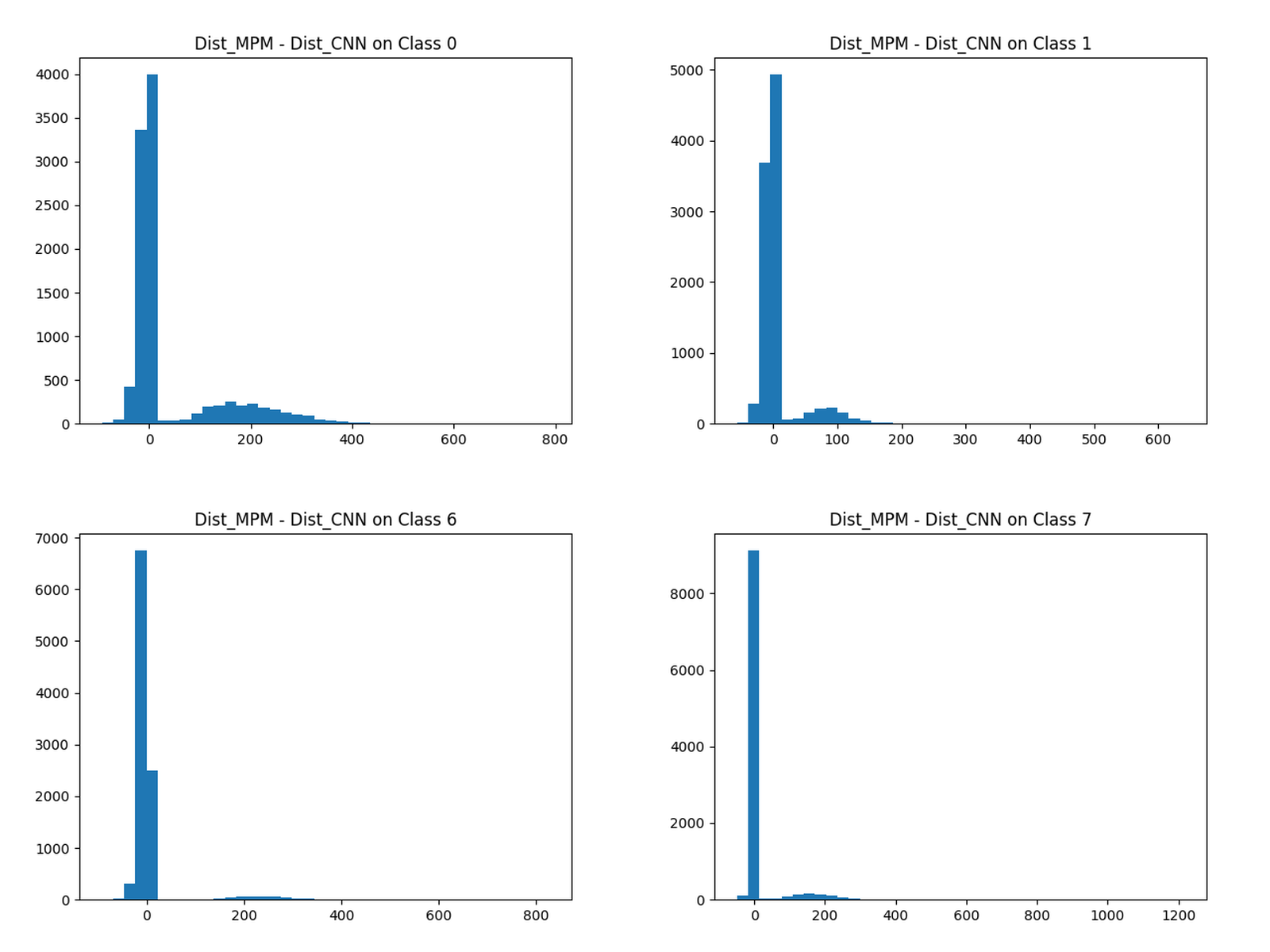}
		\vspace{-1ex}
		\caption{The differences in magnitudes of perturbation needed between the DeepMPM and the CNN.}
		\label{dist_cw_cifar_0}
	\end{centering}
\end{figure}

We also report the differences in magnitudes of perturbation needed between DeepMPM and CNN, calculated by $L2_{MPM} - L2_{CNN}$ where $L2_{MPM}$ represents the distance between DeepMPM attack example and original data, and $L2_{CNN}$ as well.
We only show the 4 classes of self-attacks on Cifar-10, as shown in Figure ~\ref{dist_cw_cifar_0}. 

From Figure ~\ref{dist_cw_cifar_0}, we could see magnitudes of perturbation needed for DeepMPM is rather larger than that of CNN, meaning the  difficulty of attacking DeepMPM is greater, proving the robustness of our model.

\section{Conclusion}
\label{sec:conclusion}
In order to alleviate the vulnerability of deep neural networks to adversarial attacks, we have proposed the Deep Minimax Probability Machine (DeepMPM), applying MPM to DNNs in an end-to-end fashion. Specifically, we put MPM on top of a deep neural network, and instead of maximizing the likelihood of labels for data, we employ the objective function of MPM. DeepMPM is more robust intuitively since it takes the global information into account and learns the worst-case bound on the probability of misclassification of future data. In practice, to evaluate the robustness of our proposed model, two groups of tasks are performed on our DeepMPM including classification and adversarial attacks on MNIST and Cifar-10.
Experimental results on classification show DeepMPM have a competitive performance with CNN. The robustness of our model has also been fully demonstrated in two attacks, the FSGM and the L2 optimization attack of C\&W.

\section*{Acknowledgments}
This paper was in part supported by Grants from the Natural Science Foundation of China(No. 61572111),  and two Fundamental Research Funds for the Central Universities of China (Nos.ZYGX2016Z003, ZYGX2017KYQD177),

\bibliographystyle{IEEEtran}
\bibliography{DeepMPM}

\begin{thebibliography}{10}
\providecommand{\url}[1]{#1}
\csname url@samestyle\endcsname
\providecommand{\newblock}{\relax}
\providecommand{\bibinfo}[2]{#2}
\providecommand{\BIBentrySTDinterwordspacing}{\spaceskip=0pt\relax}
\providecommand{\BIBentryALTinterwordstretchfactor}{4}
\providecommand{\BIBentryALTinterwordspacing}{\spaceskip=\fontdimen2\font plus
\BIBentryALTinterwordstretchfactor\fontdimen3\font minus
  \fontdimen4\font\relax}
\providecommand{\BIBforeignlanguage}[2]{{%
\expandafter\ifx\csname l@#1\endcsname\relax
\typeout{** WARNING: IEEEtran.bst: No hyphenation pattern has been}%
\typeout{** loaded for the language `#1'. Using the pattern for}%
\typeout{** the default language instead.}%
\else
\language=\csname l@#1\endcsname
\fi
#2}}
\providecommand{\BIBdecl}{\relax}
\BIBdecl

\bibitem{krizhevsky2012imagenet}
A.~Krizhevsky, I.~Sutskever, and G.~E. Hinton, ``Imagenet classification with
  deep convolutional neural networks,'' in \emph{Advances in neural information
  processing systems}, 2012, pp. 1097--1105.

\bibitem{hinton2012deep}
G.~Hinton, L.~Deng, D.~Yu, G.~E. Dahl, A.-r. Mohamed, N.~Jaitly, A.~Senior,
  V.~Vanhoucke, P.~Nguyen, T.~N. Sainath \emph{et~al.}, ``Deep neural networks
  for acoustic modeling in speech recognition: The shared views of four
  research groups,'' \emph{IEEE Signal processing magazine}, vol.~29, no.~6,
  pp. 82--97, 2012.

\bibitem{saon2015ibm}
G.~Saon, H.-K.~J. Kuo, S.~Rennie, and M.~Picheny, ``The ibm 2015 english
  conversational telephone speech recognition system,'' \emph{arXiv preprint
  arXiv:1505.05899}, 2015.

\bibitem{kalchbrenner2013recurrent}
N.~Kalchbrenner and P.~Blunsom, ``Recurrent continuous translation models,'' in
  \emph{Proceedings of the 2013 Conference on Empirical Methods in Natural
  Language Processing}, 2013, pp. 1700--1709.

\bibitem{sutskever2014sequence}
I.~Sutskever, O.~Vinyals, and Q.~V. Le, ``Sequence to sequence learning with
  neural networks,'' in \emph{Advances in neural information processing
  systems}, 2014, pp. 3104--3112.

\bibitem{szegedy2013intriguing}
C.~Szegedy, W.~Zaremba, I.~Sutskever, J.~Bruna, D.~Erhan, I.~Goodfellow, and
  R.~Fergus, ``Intriguing properties of neural networks,'' \emph{arXiv preprint
  arXiv:1312.6199}, 2013.

\bibitem{goodfellow6572explaining}
I.~J. Goodfellow, J.~Shlens, and C.~Szegedy, ``Explaining and harnessing
  adversarial examples (2014),'' \emph{arXiv preprint arXiv:1412.6572}.

\bibitem{lyu2015unified}
C.~Lyu, K.~Huang, and H.-N. Liang, ``A unified gradient regularization family
  for adversarial examples,'' in \emph{Data Mining (ICDM), 2015 IEEE
  International Conference on}.\hskip 1em plus 0.5em minus 0.4em\relax IEEE,
  2015, pp. 301--309.

\bibitem{bojarski2016end}
M.~Bojarski, D.~Del~Testa, D.~Dworakowski, B.~Firner, B.~Flepp, P.~Goyal, L.~D.
  Jackel, M.~Monfort, U.~Muller, J.~Zhang \emph{et~al.}, ``End to end learning
  for self-driving cars,'' \emph{arXiv preprint arXiv:1604.07316}, 2016.

\bibitem{maharaj2015improving}
A.~V. Maharaj, ``Improving the adversarial robustness of convnets by reduction
  of input dimensionality,'' 2015.

\bibitem{bradshaw2017adversarial}
J.~Bradshaw, A.~G. d.~G. Matthews, and Z.~Ghahramani, ``Adversarial examples,
  uncertainty, and transfer testing robustness in gaussian process hybrid deep
  networks,'' \emph{arXiv preprint arXiv:1707.02476}, 2017.

\bibitem{miyato2018virtual}
T.~Miyato, S.-i. Maeda, S.~Ishii, and M.~Koyama, ``Virtual adversarial
  training: a regularization method for supervised and semi-supervised
  learning,'' \emph{IEEE transactions on pattern analysis and machine
  intelligence}, 2018.

\bibitem{lanckriet2002minimax}
G.~Lanckriet, L.~E. Ghaoui, C.~Bhattacharyya, and M.~I. Jordan, ``Minimax
  probability machine,'' in \emph{Advances in neural information processing
  systems}, 2002, pp. 801--807.

\bibitem{huang2004minimum}
K.~Huang, H.~Yang, I.~King, M.~R. Lyu, and L.~Chan, ``The minimum error minimax
  probability machine,'' \emph{Journal of Machine Learning Research}, vol.~5,
  no. Oct, pp. 1253--1286, 2004.

\bibitem{huang2008maxi}
K.~Huang, H.~Yang, I.~King, and M.~R. Lyu, ``Maxi--min margin machine: learning
  large margin classifiers locally and globally,'' \emph{IEEE Transactions on
  Neural Networks}, vol.~19, no.~2, pp. 260--272, 2008.

\bibitem{huang2006imbalanced}
------, ``Imbalanced learning with a biased minimax probability machine,''
  \emph{IEEE Transactions on Systems, Man, and Cybernetics, Part B
  (Cybernetics)}, vol.~36, no.~4, pp. 913--923, 2006.

\bibitem{xu2007feature}
Z.~Xu, I.~King, and M.~R. Lyu, ``Feature selection based on minimum error
  minimax probability machine,'' \emph{International Journal of Pattern
  Recognition and Artificial Intelligence}, vol.~21, no.~08, pp. 1279--1292,
  2007.

\bibitem{Isii1962}
K.~Isii, ``On sharpness of tchebycheff-type inequalities,'' \emph{Annals of the
  Institute of Statistical Mathematics}, vol.~14, no.~1, pp. 185--197, Dec
  1962.

\bibitem{goodfellow2014explaining}
I.~J. Goodfellow, J.~Shlens, and C.~Szegedy, ``Explaining and harnessing
  adversarial examples,'' \emph{arXiv preprint arXiv:1412.6572}, 2014.

\bibitem{huang2015learning}
R.~Huang, B.~Xu, D.~Schuurmans, and C.~Szepesv{\'a}ri, ``Learning with a strong
  adversary,'' \emph{arXiv preprint arXiv:1511.03034}, 2015.

\bibitem{ye2018learning}
J.~Ye, L.~Wang, G.~Li, D.~Chen, S.~Zhe, X.~Chu, and Z.~Xu, ``Learning compact
  recurrent neural networks with block-term tensor decomposition,'' in
  \emph{Proceedings of the IEEE Conference on Computer Vision and Pattern
  Recognition}, 2018, pp. 9378--9387.

\bibitem{pan2019compressing}
Y.~Pan, J.~Xu, M.~Wang, J.~Ye, F.~Wang, K.~Bai, and Z.~Xu, ``Compressing
  recurrent neural networks with tensor ring for action recognition,'' in
  \emph{Proceedings of the AAAI Conference on Artificial Intelligence},
  vol.~33, 2019, pp. 4683--4690.

\bibitem{liu2018structured}
H.~Liu, L.~He, H.~Bai, B.~Dai, K.~Bai, and Z.~Xu, ``Structured inference for
  recurrent hidden semi-markov model.'' in \emph{IJCAI}, 2018, pp. 2447--2453.

\bibitem{tramer2017ensemble}
F.~Tram{\`e}r, A.~Kurakin, N.~Papernot, I.~Goodfellow, D.~Boneh, and
  P.~McDaniel, ``Ensemble adversarial training: Attacks and defenses,''
  \emph{arXiv preprint arXiv:1705.07204}, 2017.

\bibitem{madry2017towards}
A.~Madry, A.~Makelov, L.~Schmidt, D.~Tsipras, and A.~Vladu, ``Towards deep
  learning models resistant to adversarial attacks,'' \emph{arXiv preprint
  arXiv:1706.06083}, 2017.

\bibitem{papernot2016distillation}
N.~Papernot, P.~McDaniel, X.~Wu, S.~Jha, and A.~Swami, ``Distillation as a
  defense to adversarial perturbations against deep neural networks,'' in
  \emph{2016 IEEE Symposium on Security and Privacy (SP)}.\hskip 1em plus 0.5em
  minus 0.4em\relax IEEE, 2016, pp. 582--597.

\bibitem{papernot2017extending}
N.~Papernot and P.~McDaniel, ``Extending defensive distillation,'' \emph{arXiv
  preprint arXiv:1705.05264}, 2017.

\bibitem{ross2018improving}
A.~S. Ross and F.~Doshi-Velez, ``Improving the adversarial robustness and
  interpretability of deep neural networks by regularizing their input
  gradients,'' in \emph{Thirty-Second AAAI Conference on Artificial
  Intelligence}, 2018.

\bibitem{kurakin2016adversarial}
A.~Kurakin, I.~Goodfellow, and S.~Bengio, ``Adversarial machine learning at
  scale,'' \emph{arXiv preprint arXiv:1611.01236}, 2016.

\bibitem{guo2018sparse}
Y.~Guo, C.~Zhang, C.~Zhang, and Y.~Chen, ``Sparse dnns with improved
  adversarial robustness,'' in \emph{Advances in Neural Information Processing
  Systems}, 2018, pp. 240--249.

\bibitem{bertsimas2000moment}
D.~Bertsimas and J.~Sethuraman, ``Moment problems and semidefinite
  optimization,'' in \emph{Handbook of semidefinite programming}.\hskip 1em
  plus 0.5em minus 0.4em\relax Springer, 2000, pp. 469--509.

\bibitem{nesterov1994interior}
Y.~Nesterov and A.~Nemirovskii, \emph{Interior-point polynomial algorithms in
  convex programming}.\hskip 1em plus 0.5em minus 0.4em\relax Siam, 1994,
  vol.~13.

\bibitem{wang2015deep}
W.~Wang, R.~Arora, K.~Livescu, and J.~Bilmes, ``On deep multi-view
  representation learning,'' in \emph{International Conference on Machine
  Learning}, 2015, pp. 1083--1092.

\bibitem{wang2015unsupervised}
W.~Wang, R.~Arora, K.~Livescu, and J.~A. Bilmes, ``Unsupervised learning of
  acoustic features via deep canonical correlation analysis,'' in
  \emph{Acoustics, Speech and Signal Processing (ICASSP), 2015 IEEE
  International Conference on}.\hskip 1em plus 0.5em minus 0.4em\relax IEEE,
  2015, pp. 4590--4594.

\bibitem{lecun1998mnist}
Y.~LeCun, ``The mnist database of handwritten digits,'' \emph{http://yann.
  lecun. com/exdb/mnist/}, 1998.

\bibitem{krizhevsky2009learning}
A.~Krizhevsky and G.~Hinton, ``Learning multiple layers of features from tiny
  images,'' Citeseer, Tech. Rep., 2009.

\bibitem{moosavi2017universal}
S.-M. Moosavi-Dezfooli, A.~Fawzi, O.~Fawzi, and P.~Frossard, ``Universal
  adversarial perturbations,'' \emph{arXiv preprint}, 2017.

\bibitem{carlini2017towards}
N.~Carlini and D.~Wagner, ``Towards evaluating the robustness of neural
  networks,'' in \emph{2017 IEEE Symposium on Security and Privacy (SP)}.\hskip
  1em plus 0.5em minus 0.4em\relax IEEE, 2017, pp. 39--57.

\end{thebibliography}

\end{document}